\documentclass[11pt,a4paper]{article}
\usepackage{times,latexsym}
\usepackage{url}
\usepackage[T1]{fontenc}
\usepackage{booktabs}
\usepackage{float}
\usepackage{multirow}   % for \multirow
\usepackage[utf8]{inputenc}
\usepackage{arabtex}
\usepackage{utf8}
\setcode{utf8}
\usepackage{amsmath}
\usepackage{svg} % Added to support .svg images
\usepackage{amssymb}
\usepackage{graphicx}
\usepackage{amssymb}
\usepackage{colortbl}
\usepackage{xcolor}
\usepackage{paralist}
\usepackage[acceptedWithA]{tacl2021v1}
\usepackage{xspace,mfirstuc,tabulary}

\newif\iftaclinstructions
\taclinstructionsfalse % AUTHORS: do NOT set this to true
\iftaclinstructions
\renewcommand{\confidential}{}
\renewcommand{\anonsubtext}{(No author info supplied here, for consistency with
TACL-submission anonymization requirements)}
\newcommand{\instr}
\fi

\iftaclpubformat % this "if" is set by the choice of options

\else

\fi

\title{Multilingual Extraction and Recognition of Implicit Discourse Relations in Speech and Text} %\TaclPapers \\
\author{Ahmed Ruby$^{1}$~~~~Christian Hardmeier$^2$~~~~Sara Stymne$^{1}$\\[2mm]
$^1$Uppsala University, Department of Linguistics and Philology\\
$^2$IT University of Copenhagen, Department of Computer Science\\
 \texttt{\{ahmed.ruby, sara.stymne\}@lingfil.uu.se, chrha@itu.dk}}

\date{}

\begin{document}
\maketitle
\begin{abstract}
  Implicit discourse relation classification is a challenging task, as it requires inferring meaning from context. While contextual cues can be distributed across modalities and vary across languages, they are not always captured by text alone. To address this, we introduce an automatic method for distantly related and unrelated language pairs to construct a multilingual and multimodal dataset for implicit discourse relations in English, French, and Spanish.
  %by adapting an automatic method originally proposed for a single language.
  For classification, we propose a multimodal approach that integrates textual and acoustic information through Qwen2-Audio, allowing joint modeling of text and audio for implicit discourse relation classification across languages. We find that while text-based models outperform audio-based models, integrating both modalities can enhance performance, and cross-lingual transfer can provide substantial improvements for low-resource languages.
\end{abstract}

\section{Introduction}

Coherence in communication is the result of a multimodal process where relational cues are distributed across articulators and modalities, and this distribution determines the degree of overt lexical marking required \citep{HollerLevinson2019, ScholmanLaparle2025_gesturalDM}.
Discourse relations vary in their degree of explicitness across languages; some languages mark them implicitly, relying on context, intonation, and shared cultural knowledge, whereas others favor explicit markers. Accordingly, translators often add, omit, or recast discourse connectives to fit target-language norms \citep{yung-etal-2023-investigating}. Prior work by \citet{ruby-etal-2025-multimodal} has shown a tendency toward explicitation of discourse connectives in translations between two closely related varieties of the Arabic language. However, they did not examine distantly related or unrelated language pairs to highlight the challenge. We extend this line of work by investigating discourse connective explicitation across distantly related and unrelated language pairs. We adapt the pipeline proposed by \citet{ruby-etal-2025-multimodal}, which we refer to as MM-IDR (Multimodal and Multilingual for Implicit Discourse Relations), to automatically construct three datasets of paired text and speech examples from TED talks for English, French, and Spanish using translations for several languages. Table~\ref{examples} shows examples of discourse connective explicitation found in translations from the English, French, and Spanish TED talks.
Notably, translators' choices to explicate implicit discourse relations are influenced by communicative conventions such as translation norms, stylistic preferences of the target audience, and register requirements \citep{Becher2011}. Therefore, explicitation is not consistently present in all translations.

\begin{table*}[t] 
\centering
 \includegraphics[width=\textwidth]{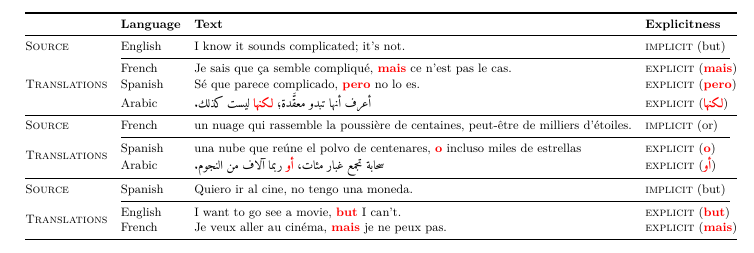}
    \caption{Examples of implicit discourse relations from TED talks (EN/FR/ES). Added connectives in translations by TED translators are highlighted.}
\label{examples}
\end{table*}

We propose a multimodal approach for implicit discourse relation (IDR) classification that leverages both textual and acoustic information for IDRs by fusing span-pooled text from a LoRA-tuned \texttt{Qwen2-Audio} \citep{xu2024qwen2audio} backbone, which jointly encodes tokenized text and its paired log–mel audio features, with compact Audio pooling statistics (conv+masked pooling on log–mels), and token-aligned prosody. We present experiments across modality settings—text-only, audio-only, and text+audio—in both monolingual and multilingual settings. To assess the contribution of each input, we perform ablations that remove audio, remove prosody, and mask text. We also report baseline results from prior work and compare them with our multimodal model. We find that text-based models consistently outperform audio-only models, but integrating text with acoustic features, especially with prosody and pooled audio representations, can further enhance performance across languages.

Our main contributions are:

\begin{compactenum}
    \item We introduce MM-IDR, a method for constructing multilingual and multimodal datasets for implicit discourse relations across distantly related and unrelated language pairs.
    \item We introduce new multimodal (text and audio) datasets for implicit discourse relations in English, French, and Spanish — the first multilingual multimodal resource for IDR.
    \item We propose a multimodal modeling approach based on an audio-language model (Qwen2-Audio), addressing limitations in its token-level representations by enriching argument spans with explicit prosodic and acoustic features.
    \item We present comprehensive experiments across monolingual and multilingual settings, including ablation studies that isolate the impact of text, audio, and prosody, showing that while text-based models consistently outperform audio-based models, integrating both modalities can enhance performance, and the multilingual setting can provide significant improvements for low-resource languages.
    
\end{compactenum}

\section{Background}

\subsection{Explicitation vs. Implicitation in languages}

Language is a fundamental way of communication that adapts to diverse cultural contexts and communicative modes, spoken or written, where cultural norms and situational demands shape how explicit or implicit meaning is conveyed \citep{Fedorenko2024Language, Nasution2022Language, brown1987politeness, gumperz1982discourse}. This adaptability becomes particularly evident when considering high-context and low-context cultures, where the level of context serves as a means of handling information overload \citep{hall1976beyond}. 

In order to identify implicit connectives in spoken languages, a key question is which languages function as high‑context systems and tend to have implicit connectives? According to Hall’s foundational work \citep{hall1976beyond} and its extensions by \citet{gudykunst1988culture}, Japanese, Chinese, and Arabic show high-context communication systems, frequently relying on implicit cues and shared cultural knowledge, while English, French, and Spanish typically function as low-context languages, though they still employ implicit communicative strategies in spoken form \citep{halliday1976cohesion, bublitz1999coherence, samovar2010communication}.

\subsection{Explicitation of connectives in translation}

Written language cannot rely on a shared here-and-now, it refines ideas into clarity and coherence—a process driven by grammatical norms, structural gaps, or cultural preferences for explicitness, necessitating explicit logical markers (e.g., ’because’, ’therefore’) in formal or written contexts. This need is particularly pronounced in written languages like Arabic, German, French, and Mandarin Chinese, where analytic grammatical structures or cultural values favoring linear logic demand overt connectives to signal relationships between ideas. For example, Arabic relies heavily on connectives such as wa (‘and’), thumma (‘then’), and fa (‘so/therefore’) to sequence ideas and signal logical relationships in formal texts \citep{ExplicitationAra, elnashar2016explicitation}, and German relies on subordinating conjunctions like weil (‘because’), obwohl (‘although’), and adverbs like deshalb (‘therefore’) to structure complex arguments in formal writing \citep{stede-2008-connective, durrell2015_german}. French, Spanish, and Italian rely on adverbs like ainsi (‘thus’), por lo tanto (‘therefore’), and quindi (‘so’) to structure complex arguments \citep{hawkins2001french, kattan2003spanish, proudfoot2005italian}. Mandarin Chinese employs correlative conjunctions like (‘because…therefore’) to express cause and effect, reflecting grammatical necessity and classical tradition \cite{Harlow1992Chinese}.

Ultimately, the degree of explicitness reflects a language’s grammatical architecture and cultural priorities. For example, English employs less semantically specific connectives than German \citep{yung-etal-2023-investigating, HOEK2017113}, while Arabic is more explicitative than English due to its verbal grammatical structures \citep{elnashar2016explicitation}.

\section{Related Work}

Implicit discourse relation recognition has long been recognized as one of the most challenging tasks in NLP \citep{kurfali-ostling-2021-lets, kim-etal-2020-implicit, Xiang2023IDRSurvey}. This difficulty arises from the absence of explicit connectives, requiring models to infer relations from semantics and broader context. Most existing datasets for implicit discourse relations work are monolingual and text-based. The Penn Discourse Treebank (PDTB) \citep{prasad2008penn, prasad2019penn}, the largest resource for English, reveals the extent of this challenge, as implicit relations do not achieve accuracy comparable to explicit ones. Similar monolingual resources exist for other languages \citep{long-etal-2020-ted, turkishdataset, oza-etal-2009-hindi, prasertsom-etal-2024-thai, ogrodniczuk-etal-2024-polish-discourse}, while multilingual datasets such as TED-MDB \citep{zeyrek2019ted} and DiscoGeM \citep{scholman-etal-2022-discogem, yung-etal-2024-discogem} cover multiple languages—but all remain text-only.

There have been efforts to explore using prosodic cues and audio features along with text in implicit discourse relations identification \citep{murray-etal-2006-prosodic, Kleinhans2017, ruby24_speechprosody}, showing a promising direction, and recently \citet{ruby-etal-2025-multimodal} constructed a text-audio aligned Arabic discourse dataset using two variants of the same language (Egyptian Arabic and Modern Standard Arabic (MSA)) to identify implicit-to-explicit connective pairs. Yet multimodal discourse datasets remain limited to single languages, and the potential of LLMs for audio-enhanced discourse relations remains underexplored.

\section{Multilingual Dataset Construction}

\begin{figure*}[t]
    %\vspace{-1.6cm} %
    \noindent %
    \centering
    \includegraphics[width=0.9\textwidth]{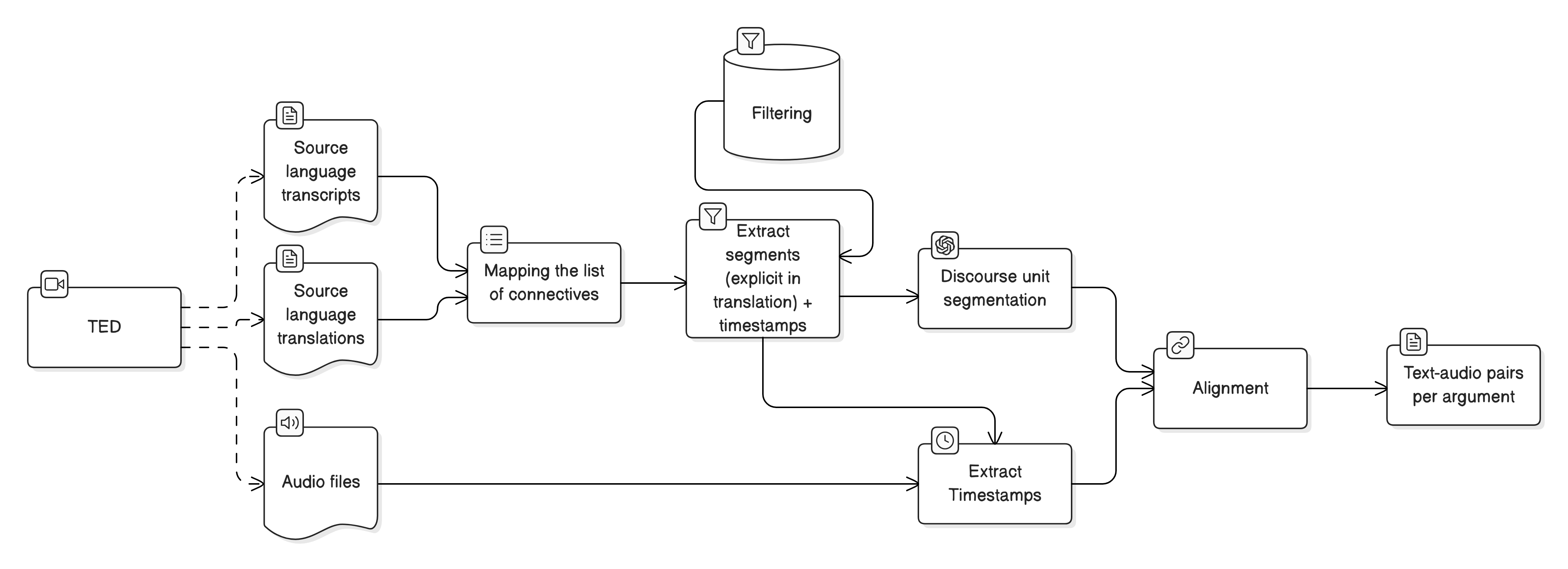}
 \caption {Pipeline for extracting multimodal implicit discourse relations in: English, French, and Spanish.}
\label{fig:flowchart}
\end{figure*}

To construct datasets for implicit discourse relations in distantly related language pairs, we adapt the pipeline proposed by \citet{ruby-etal-2025-multimodal}, which extracts semantically equivalent pairs of implicit and explicit discourse connectives by aligning speech transcripts in Egyptian Arabic with their corresponding subtitles in MSA (both variants of the same language), as shown in Figure \ref{fig:flowchart}, with several modifications described in Section~\ref{sec:mods}.

\subsection{Data Collection and Preparation} 

We collected 729 multilingual TEDx talks across three languages—English, French, and Spanish—each accompanied by transcriptions. 

We focused on five languages: Arabic, English, French, German, and Spanish. The number of available translations varies per talk, some include all four languages, others fewer.
Table~\ref{tab:ted-nots} shows an overview of the multilingual TEDx talks with their available translations.

\begin{table}[t!]
\centering
\begin{tabular}{@{}lcc@{}}
\toprule
\textbf{Language} & \textbf{TED\_Talks} & \textbf{Translations} \\
\midrule
English   & 348 & ar, es, fr, de \\
French    & 217 & ar, en, es \\
Spanish   & 164 & ar, en, fr \\
\bottomrule
\end{tabular}
\caption{The number of TED talks for each language and their available translations in target languages.}
\label{tab:ted-nots}
\end{table}

\subsection{Implicit-to-Explicit Connective Mapping}

In order to identify implicit connectives through their explicitation across multiple language pairs, we first prepared language-specific connective lists. 
Following \citet{ruby-etal-2025-multimodal}'s approach to MSA, we use Connective Lex\footnote{\url{http://connective-lex.info/}} \citep{stede2019connectivelex} to extract unambiguous discourse connectives in 
English, French, German, and Spanish. We then map these connectives across the directional pairs (e.g., EN$\rightarrow$\{AR, FR, ES, DE\}) to identify implicit-to-explicit instances. We defined several heuristic rules to filter out non-discourse uses of the potential connectives (e.g., "so" as a filler or an intensifier).

For discourse relation labeling, we follow \citet{ruby-etal-2025-multimodal} by using their discourse relations categories based on explicitated connectives, which aligns with PDTB-3 level 1 taxonomy \citep{wolf-gibson-2005-representing, prasad2019penn}. However, we broadened their 'Temporal Sequence' category to 'Temporal' to include non-sequential temporal relations found in our data, resulting in four final categories: Cause-Effect, Contrast, Elaboration, and Temporal.

\subsection{Segmenting Discourse Units}
\label{sec:length}

After identifying implicit-to-explicit connective mappings, we extract the source subtitle segments where implicit connectives occur, along with their surrounding context and timestamps. For each identified segment with an implicit connective, we extract three consecutive sentences from the source using punctuation boundaries: the sentence containing the implicit relation, the sentence before it, and the sentence after it. We mark the boundaries of the relevant segment with special markers to guide the segmentation process.
We use OpenAI’s GPT-4o \citep{openai2024gpt40}, in a few-shot setting, to process this annotated context and identify Arg1 and Arg2, following PDTB minimal span principles. A small human evaluation (see Section~\ref{sec:quality}) shows the spans are reliable.

\subsection{Aligning Discourse Units} 

To obtain precise audio boundaries for Arg1 and Arg2, we extract the three-sentence context audio using subtitle timestamps. We process this audio through Faster-Whisper ASR with word-level timestamps, and then match Arg1 and Arg2 spans to these timestamped words to identify precise start and end times for each span. Using those boundaries, we then extract the Arg1 and Arg2 audio segments, yielding a parallel text-audio discourse dataset. 

\subsection{Modifications to the Original Pipeline}
\label{sec:mods}
We made the following modifications to adapt the pipeline by \citet{ruby-etal-2025-multimodal} for constructing multilingual datasets:

\begin{compactitem}

\item Since different languages explicitate different connectives, we extend implicit-to-explicit connective mapping to multiple target languages to capture a broader range of implicit connectives. Using three source languages, we map each to four target languages, creating the following 
pairs: EN$\rightarrow$\{AR, FR, ES, DE\}, FR$\rightarrow$\{AR, EN, ES, DE\}, and ES$\rightarrow$\{AR, EN, FR, DE\}, depending on translation availability for each TED talk.

\item Unlike the original pipeline, which required ASR to transcribe the actual Egyptian Arabic speech (since available subtitles were in MSA), we work with languages where TED provides high-quality human transcriptions that directly match the spoken language. We therefore use these transcriptions directly, eliminating the ASR step—a straightforward simplification and improvement of the pipeline.

\item While the original pipeline used GPT-4 \citep{openai2023gpt4} with LADTB \citep{AlSaif2012} guidelines to perform discourse unit segmentation on MSA text before aligning to Egyptian Arabic, we prompt GPT-4o \citep{openai2024gpt40} with few-shot examples and instructions to follow PDTB \citep{prasad2019penn} segmentation principles, specifically the minimal span style. We apply this approach consistently across all three languages.

\item We extended the 'Temporal Sequence' category to 'Temporal' to include non-sequential temporal relations found in our data.
\item We add a filtering step to accurately capture implicit connectives when mapping between source and target texts by: 

\begin{compactitem}
    \item validating segment pairs using timestamp duration consistency to ensure that aligned segments actually correspond to each other before mapping implicit-to-explicit connectives.
    \item validating the potential implicit connective cases by screening both source and target segments (not just initial positions) for pre-existing connectives or alternative connectives that would indicate explicit-to-explicit translation.
    \item defining several heuristic rules to filter out non-discourse uses of the potential connectives. 
    \item removing duplicate examples that arise from mapping each source to multiple targets %(e.g., EN$\rightarrow$AR, EN$\rightarrow$ES, EN$\rightarrow$FR),
    as the same implicit connective may be explicitated across translations.
    
\end{compactitem}
\item We use Faster-Whisper 
\citep{fasterwhisper2023} instead of Whisper-timestamped \citep{lintoai2023whispertimestamped}
for Arg1/\allowbreak Arg2 extraction from the audio,  as it provides more accurate word-level timestamps that are needed to precisely cut argument boundaries.

\end{compactitem}

\subsection{Quality Control}
\label{sec:quality}

We adopted the automatic pipeline of \citet{ruby-etal-2025-multimodal}, which was shown to give high-quality data. They checked all 760 Arabic instances, and found segmentation errors only in 2.5\% of the examples and 1.4\% of instances where the examples had no discourse function. There were no cases of mismatched discourse labels. We expect a similar accuracy in the new datasets, and conducted a small human evaluation of 100 samples, verified by two persons.
We found no instances of mismatched discourse labels. We identified 6 segmentation errors, in all cases overlapping with the correct segment: 4 examples that included extraneous content (such as parenthetical phrases or unrelated subsequent clauses) in Arg1 or Arg2, and 2 examples where the boundary of Arg2 was cut too early, failing to capture the complete meaning of Arg2. In addition, we found 2 cases that were incorrectly identified as implicit, where the system had failed to filter examples that contained the explicit connective ``instead'', meaning that the examples were not implicit. This could be filtered in future dataset releases. In summary, the resource is of high-quality, even though the segmentation in a few examples, 6\% in our sample, can fail to cover the complete span, or include extra material.

\subsection{Dataset Statistics and Splits}

We split the datasets into three distinct sets with no overlap in talks between sets, as summarized in Table~\ref{distribution}. For English, we used a 60\%/20\%/20\% split (training/validation/testing). For French, following the Arabic dataset's \citep{ruby-etal-2025-multimodal} distribution due to their similar sizes, we allocated 55\%/15\%/30\% (training/validation/testing). For Spanish, given its relatively small size, we prioritized test data with a 25\%/25\%/50\% split (training/validation/testing) to ensure robust evaluation. These splits maintain consistent class distributions across sets. Table~\ref{relation_distribution} presents the class distribution of relations across languages.

\begin{table}
\small
\centering
\begin{tabular}{llll}
\hline {Language} & {Dataset} & {Relations} & {Talks}\\ \hline
\multirow{3}{*}{English} & Train       & 1563 & 188 \\
                         & Validation  &  520 &  78 \\
                         & Test        &  520 &  82 \\ \hline
\multirow{3}{*}{French}  & Train       &  457 & 130 \\
                         & Validation  &  124 &  29 \\
                         & Test        &  249 &  58 \\ \hline
\multirow{3}{*}{Spanish} & Train       &  101 &  49 \\
                         & Validation  &  100 &  33 \\
                         & Test        &  201 &  82 \\\hline

\multirow{3}{*}{\textit{Arabic}} & \textit{Train}       &  418 &  43 \\
                         & \textit{Validation}  &  \textit{114} &  \textit{17} \\
                         & \textit{Test}        &  \textit{228} &  \textit{27} \\
          
\hline
\end{tabular}
\caption{\label{distribution}Multilingual dataset distribution: implicit discourse relations and number of talks by split. The table presents our newly constructed datasets for English, French, and Spanish, along with an existing Arabic dataset \citep{ruby-etal-2025-multimodal} included for comparison.}
\end{table}

\begin{table}
\small
\centering
\resizebox{\columnwidth}{!}{%
\begin{tabular}{@{}lccccc@{}}
\toprule
\multicolumn{1}{c}{\multirow[c]{2}{*}{\textbf{Relation type}}}
  & \multicolumn{3}{c}{\textbf{Languages}} \\ 
\cmidrule(l){2-5}
 & English & Spanish & French & \textit{Arabic}\\ 
\midrule
’cause-effect’  & 593 & 97 & 169 & \textit{216}\\
’contrast’   & 704 & 101 & 199 & \textit{212}\\
’temporal’    & 546 & 73 & 127 & \textit{135}\\
’elaboration’      & 760 & 131 & 335 & \textit{197}\\
%Photo    & X & X & X & X & X \\
%Formula  & X & X & X & X & X \\
%Table    & X & X & X & X & X \\
%Text     & X & X & X & X & X \\ 
\hline
\textbf{Total} & 2603  & 402 & 830 & \textit{760} \\
\bottomrule
\end{tabular}
}
\caption{Class distribution of implicit discourse relations across languages: number of examples per relation type. The table presents our newly constructed datasets for English, French, and Spanish, along with an existing Arabic dataset \citep{ruby-etal-2025-multimodal} included for experimental comparison.}
\label{relation_distribution}
\end{table}

\section{Explicitation vs. Human Annotation}

To investigate the coverage of our resource, we compared it with a resource for implicit discourse relations annotated by humans, the English data from TED-MDB \citep{zeyrek2019ted}. We do not expect the explicitation method to cover all instances, since it relies on translators' decisions to explicitly insert connectives in the target language translations, in contrast to human annotators, who were tasked with identifying all instances of implicit relations. We note that TED-MDB only covers inter-sentential relations, whereas our method also extracts intra-sentential relations.

In order to examine which explicitation patterns align with human-annotated discourse relations and to explore the differences, we obtained translations of  English TED-MDB talks in six languages (Arabic, Chinese, French, German, Italian, and Spanish). We then systematically identified instances where implicit English discourse relations were made explicit through connectives in the translations using our adapted MM-IDR method, and compared these cases to the annotations in TED-MDB.

Table~\ref{MM-IDR-TED-MDB} shows the comparison between human-annotated implicit discourse relations in TED-MDB and automatic annotation through MM-IDR. While MM-IDR captured only 55 of the 192 TED-MDB relations (28.6\%), it identified 9 additional inter-sentential relations not present in TED-MDB, and 90 new intra-sentential relations, which were outside of TED-MDB's scope. Notably, Arabic showed the highest tendency toward explicitation, which aligns with both the literature and our findings when we constructed our dataset, followed by German and Chinese, while French and Spanish showed the lowest explicitation rates.

\begin{table}%[ht]
\centering
\resizebox{\columnwidth}{!}{%
\begin{tabular}{@{}lcccc@{}}
\small
\begin{tabular}{lccccc}
\toprule
& & \multicolumn{3}{c}{\textbf{Inter-S}} & \\
\cmidrule(lr){3-5}
\textbf{Language} & \textbf{Intra-S} & \textbf{Total} & \textbf{Matching} & \textbf{New} &  \\
\midrule
\textbf{TED-MDB (English)} & — & 192 & — & — \\
\midrule
\multicolumn{5}{l}{\textit{MM-IDR annotation in translations:}} \\
Arabic          & 30 & 42 & 39 & 3  \\
Chinese         & 17 & 10 &  8 & 2  \\
French          &  6 &  5 &  2 & 3  \\
German          & 26 & 15 &  9 & 6  \\
Italian         & 12 &  7 &  5 & 2  \\
Spanish         &  7 &  5 &  3 & 2  \\
\midrule
Total (MM-IDR) & 98 & 84 & 66 & 18  \\
\midrule
\textbf{Unique Connectives (All)} & \textbf{90} & \textbf{64} & \textbf{55} & \textbf{9} \\
Shared ($\geq$ 2 Languages) & 8 & 10 & — & — \\
\bottomrule
\end{tabular}
\end{tabular}
}
\caption{Annotation statistics for implicit discourse relations in the English TED-MDB corpus: human TED-MDB annotations vs. MM-IDR automatic annotations from translation-based connective explicitation across six languages. The table presents both intra-sentential relations (not annotated in TED-MDB) and inter-sentential relations across annotations.}
\label{MM-IDR-TED-MDB}
\end{table}

\section{Implicit Discourse Relation Modeling}

In this section we present methods for implicit discourse relation classification, including a set of baselines, and our proposed new method.

\subsection{Baselines}
We selected the two best performing baseline methods for each modality from \citet{ruby-etal-2025-multimodal}.
For text, we chose BERT (replacing AraBERT \citep{antoun-etal-2020-arabert} with BERT-base for English \citep{devlin-etal-2019-bert}, CamemBERT-large for French \citep{martin-etal-2020-camembert}, and BETO-base for Spanish \citep{CaneteCFP2020}) and TF-IDF + Logistic Regression (LogReg). For audio, we selected Prosodic+LogReg and Whisper \citep{whisperopenai} + Neural Network (NN). For text-audio, we used BERT+Wav2vec2 \citep{Baevski2020wav2vec2A}, which combines each language-specific BERT model's [CLS] with Wav2vec speech features, and Prosodic + TF-IDF+LogReg.

\subsection{Proposed Model}

\begin{figure*}
    %\vspace{-1.6cm} %
    \noindent %
    \centering
    \includegraphics[width=1.\textwidth]{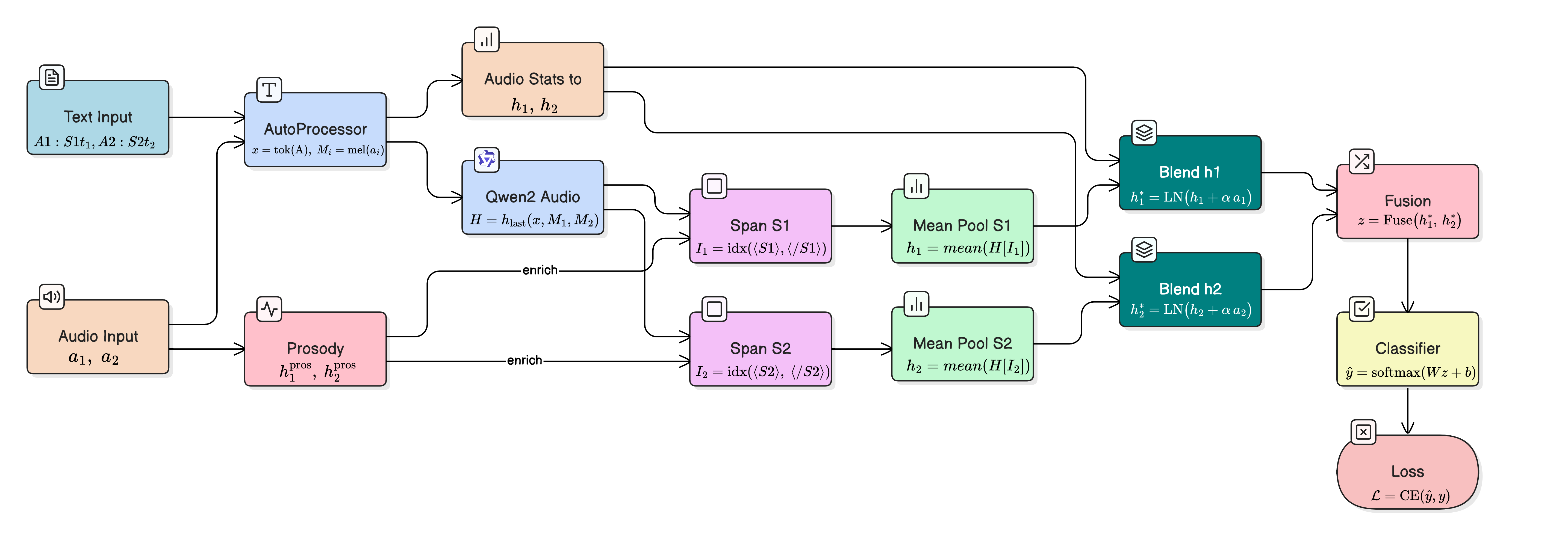}
 \caption {Architecture for our proposed model for implicit discourse relation classification.}
\label{fig:Arch}
\end{figure*}

We introduce a multimodal framework for implicit discourse relation classification that jointly leverages text and speech, as shown in Figure \ref{fig:Arch}. Each instance contains two text segments $(t_1,t_2)$, corresponding to Arg1 and Arg2, with aligned audio waveforms $(a_1,a_2)$. To enable precise argument boundary detection for pooling and analysis, we mark the spans explicitly using \texttt{<S1>}\dots\texttt{</S1>} and \texttt{<S2>}\dots\texttt{</S2>} \citep{baldini-soares-etal-2019-matching, dai-huang-2018-improving, xu2024qwen2audio}. A pretrained audio–language model (Qwen2-Audio \citep{xu2024qwen2audio}) processes the tokenized text and the two audio segments (log-mel features), producing token-level hidden states $H_i\in \mathbb{R}^{T \times d}$. For each argument, we mean pool the hidden states between its markers to obtain argument-level vectors $h_1$ and $h_2$.

\paragraph{Prosody-enhanced spans}
To enrich the text spans with speech cues that might not be well represented in log–mel spectrograms, we extract word-aligned prosodic descriptors from $(a_1,a_2)$ using Faster-Whisper timestamps to obtain word timestamps. For each argument $i$, we compute a per-word prosody sequence $P_i \in \mathbb{R}^{W_i \times D_p}$ summarizing pitch, energy, and timing information. We project $P_i$ to match the hidden state dimension of $H$ and fuse it into the corresponding token span via cross-attention before pooling. For each token $h_t$ in $H_{\text{span},i}$, we compute attention scores over the projected prosody vectors $\hat{P}_i$ to measure how relevant each prosody vector is to that token, normalize these scores into weights via softmax, and take a weighted average of the prosody vectors. Each token thus receives a prosody summary relevant to it, which is scaled by a learnable weight $\gamma$ and added back to the original representation (with layer normalization), and then mean-pooled to obtain $h_i$:
\vspace{-1mm}
\begin{gather}
\tilde{H}_i = \operatorname{LN}\!\Big(H_{\text{span},i} + \gamma\,\operatorname{Attn}\!\big(H_{\text{span},i},\,\hat{P}_i\big)\Big), \\
h_i = \operatorname{pool}(\tilde{H}_i)
\end{gather}

where $\mathrm{Attn}(Q, K)$ denotes cross-attention with queries $Q$ and keys/values $K$.

\paragraph{Audio pooling statistics}
While Qwen2-Audio integrates audio into token-level representations, it does not produce an explicit argument-level acoustic summary, and global or time-varying cues can be averaged out by mean-pooling over token spans. Following \cite{okabe18_interspeech}, we apply statistics pooling over log-mel frames, using a padding mask so that only real frames contribute. For each segment, we pass the log-mel frames through two 1D convolutional layers, compute mean and standard deviation over non-padded frames of the convolved output, and project to match the hidden state dimension of $H$, yielding audio vectors $a_1$ and $a_2$. The audio vector is scaled by a small fixed weight $\alpha$ and added to the corresponding argument vector:

\[
\tilde{h}_i = \mathrm{LN}\!\big(h_i + \alpha \, a_i\big)
\]

\paragraph{Classification} 

Following \cite{guo-etal-2018-implicit, ruan-etal-2020-interactively}, we apply bidirectional cross-attention fusion over $(\tilde{h}_1, \tilde{h}_2)$: $\tilde{h}_1$ attends to $\tilde{h}_2$ and $\tilde{h}_2$ attends to $\tilde{h}_1$. We concatenate the two outputs, pass them through an MLP, and apply $\ell_2$ normalization to obtain $z$, which is passed to a two-layer classifier for relation prediction.

\paragraph{Implementation}
For training, we fine-tune \texttt{Qwen2-Audio-7B-Instruct} \citep{xu2024qwen2audio}, a pretrained audio-language model backbone, using LoRA adapters under 4-bit quantization for efficiency, while training the task-specific heads end-to-end.
The training objective is cross-entropy loss. We optimize using AdamW with a cosine schedule, 10\% warmup, weight decay 0.05, and max grad-norm 1.0. We use LR $5\!\times\!10^{-5}$ for the base model and text heads, and $5\!\times\!10^{-4}$ for the audio pooling statistics head. We train for 7 epochs with early stopping based on validation loss, saving each epoch and loading the best checkpoint. To address class imbalance, we apply class weights. For more detailed hyperparameters, see Appendix \ref{sec:appendix0}. For each experiment, we run the model 3 times with different random seeds and report the average scores.
We report all results using accuracy and macro-averaged precision, recall, and F1-score across all relations.

\section{Results}

We run all the experiments on the three newly constructed datasets (English, French, and Spanish), as well as on the existing Arabic dataset by \citet{ruby-etal-2025-multimodal}.

Table~\ref{baseline} shows the baseline results for implicit discourse relation classification across four languages using audio-only, text-only, and both. Text-based models consistently outperform audio-only models across all languages, with BERT achieving the highest F1 scores for English (0.53). Multimodal combinations yield mixed results; simpler models (Prosodic\allowbreak{} +\allowbreak{} TF-IDF\allowbreak{} +\allowbreak{} LogReg) show minimal or no improvements across all languages, while advanced models (BERT\allowbreak{} +\allowbreak{} Wav2vec2, BETO\allowbreak{} +\allowbreak{} Wav2vec2) improve French and Spanish but show no gain for Arabic and slight degradation for English. This substantial improvement for Spanish in multimodal fusion is likely due to limited discourse training data, which produces weaker text-only performance compared to other languages, making speech features comparatively more valuable. These results reveal that multimodal fusion, though promising in principle, requires careful consideration. When the text modality is poorly represented due to limited training data, as in Spanish, speech features can effectively bridge the performance gap. Per-class F1 scores are provided in Table~\ref{per-class1} (Appendix \ref{sec:appendix1}).

\begin{table*}
%\small
\centering
\resizebox{\textwidth}{!}{%
\begin{tabular}{c|c|cccc|cccc|cccc|cccc}
\multirow{2}{*}{Data} & \multirow{2}{*}{Models} & \multicolumn{4}{c|}{Arabic} & \multicolumn{4}{c|}{English} & \multicolumn{4}{c|}{French} & \multicolumn{4}{c}{Spanish} \\ \cline{3-18}
& & P & R & F1 & Acc & P & R & F1 & Acc & P & R & F1 & Acc & P & R & F1 & Acc \\
\hline
\multirow{2}{*}{Audio} 
& Prosodic+LogReg & 0.30 & 0.28 & 0.28 & 0.29 & 0.30 & 0.30 & 0.29 & 0.30 & 0.23 & 0.23 & 0.22 & 0.22 & 0.20 & 0.21 & 0.20 & 0.20 \\
& Whisper+NN & 0.32 & 0.31 & 0.31 & 0.33 & 0.36 & 0.36 & 0.35 & 0.38 & 0.29 & 0.30 & 0.29 & 0.34 & 0.26 & 0.25 & 0.25 & 0.25 \\
\hline
\multirow{2}{*}{Text} 
& TF-IDF+LogReg & 0.42 & 0.43 & 0.40 & 0.42 & 0.40 & 0.40 & 0.40 & 0.40 & 0.36 & 0.36 & 0.36 & 0.39 & 0.35 & 0.33 & 0.33 & 0.34 \\
& BERT & 0.42 & \textbf{0.43} & 0.41 & 0.43 & \textbf{0.55} & \textbf{0.52} & \textbf{0.53} & \textbf{0.53} & 0.50 & 0.43 & 0.42 & 0.49 & 0.23 & 0.30 & 0.24 & 0.31 \\
\hline
\multirow{2}{*}{Both} 
& Prosodic+TF-IDF+LogReg & \textbf{0.43} & \textbf{0.43} & \textbf{0.42} & \textbf{0.43} & 0.40 & 0.40 & 0.40 & 0.40 & 0.38 & 0.37 & 0.37 & 0.40 & 0.34 & 0.31 & 0.30 & 0.31 \\
& BERT+Wav2vec2 & \textbf{0.43} & 0.41 & 0.41 & 0.43 & 0.53 & 0.51 & 0.51 & 0.53 & \textbf{0.52} & \textbf{0.52} & \textbf{0.52} & \textbf{0.53} & \textbf{0.48} & \textbf{0.38} & \textbf{0.38} & \textbf{0.39} \\
\hline                   
\end{tabular}
}
\caption{\label{baseline} Test-set results for baseline methods in IDR classification across different modalities.}
\end{table*}

\begin{table*}[t]
\small
\centering
\begin{tabular}{c|c|cccc|cccc}
{Language} & \multirow{3}{*}{Data} & \multicolumn{4}{c|}{Language-Specific Training} & \multicolumn{4}{c}{Multilingual Training} \\
\cline{1-1}\cline{3-10}
 Test& & P & R & F1 & Acc. & P & R & F1 & Acc. \\
\hline

\multirow{3}{*}{Ar} & Audio 
    & 0.07 & 0.25 & 0.11 & 0.29  & 0.05 & 0.25 & 0.11 & 0.30\\ 
& Text 
    & 0.44 & 0.44 & 0.43 & 0.44 & \textbf{0.58} & \textbf{0.44} & \textbf{0.38} & \textbf{0.47}\\ 
& Both 
    & \textbf{0.48} & \textbf{0.48} & \textbf{0.48} & \textbf{0.49} & 0.53 & 0.40 & 0.36 & 0.42\\ 
\hline
\multirow{3}{*}{En} & Audio 
    & 0.07 & 0.25 & 0.12 & 0.31 & 0.07 & 0.25 & 0.12 & 0.31\\ 
& Text 
    & \textbf{0.62} & 0.52 & 0.53 & 0.54 & \textbf{0.60} & \textbf{0.52} & \textbf{0.53} & \textbf{0.55}\\ 
& Both 
    & 0.57 & \textbf{0.54} & \textbf{0.54} & \textbf{0.55} & 0.58 & 0.50 & 0.48 & 0.51 \\ 
\hline
\multirow{3}{*}{Fr} & Audio 
    & 0.05 & 0.25 & 0.09 & 0.23  & 0.08 & 0.24 & 0.12 & 0.24 \\ 
& Text 
    & 0.41 & 0.40 & 0.40 & 0.44  & \textbf{0.51} & 0.44 & 0.43 & 0.49\\ 
& Both 
    & \textbf{0.43} & \textbf{0.43} & \textbf{0.43} & \textbf{0.46}  & \textbf{0.51} & \textbf{0.48} & \textbf{0.48} & \textbf{0.50} \\ 
\hline
\multirow{3}{*}{Es} & Audio 
    & 0.07 & 0.25 & 0.12 & 0.30  & 0.06 & 0.25 & 0.11 & 0.27\\ 
& Text 
    & \textbf{0.55} & \textbf{0.49} & \textbf{0.46} & \textbf{0.47}  & 0.55 & 0.49 & 0.48 & 0.50\\ 
& Both 
    & 0.51 & 0.41 & 0.40 & 0.44  & \textbf{0.58} & \textbf{0.53} & \textbf{0.53} & \textbf{0.54} \\ 
\hline

\end{tabular}
\caption{\label{model} Test-set results for our proposed method in IDR classification across different modalities (Text, Audio, and combined Text-Audio), using monolingual training (i.e., models trained and tested on the same language), and using multilingual training (i.e., models trained on all four languages and tested individually)}.
\end{table*}

\begin{table*}[t]
\small
\footnotesize
\resizebox{\textwidth}{!}{%
\centering
\begin{tabular}{cccc|cccc|cccc|cccc|cccc}
\multicolumn{4}{c|}{Model Components} & \multicolumn{4}{c|}{Arabic} & \multicolumn{4}{c|}{English} & \multicolumn{4}{c|}{French} & \multicolumn{4}{c}{Spanish} \\
\hline
T & A & APS & Pr & P & R & F1 & Acc & P & R & F1 & Acc & P & R & F1 & Acc & P & R & F1 & Acc \\
\hline
\checkmark & \checkmark & \checkmark & \checkmark & 0.48 & 0.48 & \textbf{0.48} & \textbf{0.49} & 0.57 & \textbf{0.54} & \textbf{0.54} & \textbf{0.55} & 0.43 & 0.43 & 0.43 & 0.46 & 0.51 & 0.41 & 0.40 & 0.44 \\
\checkmark & \checkmark & \checkmark &   & \textbf{0.49} & 0.48 & \textbf{0.48} & \textbf{0.49} & 0.58 & 0.47 & 0.46 & 0.51 & \textbf{0.49} & \textbf{0.51} & \textbf{0.50} & \textbf{0.51} & 0.47 & 0.40 & 0.39 & 0.44 \\
\checkmark &   &   &   & 0.44 & 0.44 & 0.43 & 0.44 & \textbf{0.62} & 0.52 & 0.53 & 0.54 & 0.41 & 0.40 & 0.40 & 0.44 & 0.55 & \textbf{0.49} & \textbf{0.46} & \textbf{0.47} \\
\checkmark & \checkmark &   & \checkmark & 0.47 & 0.48 & 0.46 & 0.48 & 0.57 & 0.52 & 0.52 & 0.53 & 0.41 & 0.42 & 0.41 & 0.45 & 0.49 & 0.36 & 0.34 & 0.40 \\
\checkmark &   &   & \checkmark & 0.47 & 0.48 & 0.46 & 0.48 & 0.57 & 0.51 & 0.51 & 0.52 & 0.47 & 0.47 & 0.47 & 0.48 & \textbf{0.58} & 0.40 & 0.36 & 0.44 \\
\checkmark & \checkmark &   &   & 0.46 & 0.46 & 0.45 & 0.46 & 0.46 & 0.42 & 0.38 & 0.47 & 0.45 & 0.43 & 0.43 & 0.47 & 0.48 & 0.37 & 0.33 & 0.41 \\
  & \checkmark & \checkmark & \checkmark & 0.07 & 0.25 & 0.11 & 0.29 & 0.14 & 0.25 & 0.17 & 0.31 & 0.18 & 0.24 & 0.11 & 0.24 & 0.07 & 0.25 & 0.12 & 0.30 \\
  & \checkmark & \checkmark &   & 0.07 & 0.25 & 0.12 & 0.31 & 0.07 & 0.25 & 0.12 & 0.31 & 0.05 & 0.25 & 0.09 & 0.23 & 0.07 & 0.25 & 0.12 & 0.30 \\
\hline
\end{tabular}
}
\caption{\label{ablation} Test-set ablation results for our proposed method in IDR classification across different modalities (Text, Audio, and combined Text-Audio), using monolingual training (i.e., models trained and tested on the same language) T=Text, A=Audio, APS=Audio Pooling Stats, Pr=Prosody.}
\end{table*}

Table~\ref{model} shows the results of our proposed model for language-specific and multilingual training settings. For language-specific training, where models are trained and tested on the same language using three modality configurations: audio-only, text-only, and multimodal (both audio and text). Audio-only models perform poorly across all languages, highlighting the limited predictive power of speech information alone for discourse relation classification. This suggests that while the proposed architecture captures certain discourse patterns, it struggles with precision and overall discriminative power compared to established audio feature extraction methods.
Text-based models achieved significantly stronger performance than audio approaches across all languages. The proposed text-only model performed comparably to BERT-based baselines in English, Arabic, and French, indicating that these transformer-based systems had effectively captured the available textual signals for discourse relation classification. However, in Spanish, the proposed model substantially outperformed both baselines, particularly BETO, suggesting better capacity to capture discourse-level relational cues in this language even with limited training data.
Multimodal models combining audio and text yielded mixed results across languages. In Arabic and English, the full system—integrating text, audio embeddings, prosodic features, and audio pooling statistics—outperformed traditional feature-concatenation baselines, demonstrating the benefits of joint multimodal modeling when both modalities contribute meaningful information. French also showed improvement over the text-only model but still underperformed compared to the BERT + Wav2vec2 baseline, suggesting that the proposed fusion strategy was less effective for this language. In contrast, Spanish demonstrated negative fusion, where multimodal integration underperformed the text-only model, even though baseline results had shown fusion benefits under similar low-resource conditions. This indicates that the audio signals from the proposed architecture may have disrupted the strong text-based predictions in this low-resource setting.

For the multilingual training setting, where models are trained jointly on all four languages and tested separately on each, using three modality configurations: audio-only, text-only, and multimodal (both audio and text). Audio-only models remain consistently poor across all languages, showing little to no benefit from joint training. This suggests that prosodic and acoustic features alone are not sufficient, even when exposed to a larger and more diverse training set.

Text-only models in this setting yield mixed results across languages. For Arabic, performance was lower compared to monolingual training, suggesting negative transfer from other languages. In contrast, English maintains comparable performance, indicating robustness to multilingual training. Meanwhile, French and Spanish show improvements compared to their monolingual counterparts, suggesting that multilingual training can enhance generalization for some languages by leveraging cross-lingual transfer through shared discourse representations.

For multimodal fusion, models in this setting reveal language-specific patterns. For Spanish and French, performance improves when audio is combined with text, suggesting that richer cross-lingual representations help reinforce the learning signal. In contrast, for Arabic and English, the multimodal models underperform compared to their text-only counterparts, possibly due to interference from less informative audio features in the multilingual context. Overall, these results reveal that multilingual training can provide significant improvements for low-resource languages. Per-class F1 scores for both language-specific and multilingual settings are provided in Table~\ref{per-class2}  (Appendix \ref{sec:appendix2}).

\subsection{Ablation Study}

To investigate the contribution of text, raw audio, and prosody in our model, we conduct a controlled ablation study. Each ablation run uses a fresh initialization, model (4-bit base + fresh LoRA), and fresh heads, all trained with identical configurations (hyperparameters, sampler, early stopping, and seed). This ensures no parameter sharing or leakage across runs. We ablate four factors: text masking (none vs. empty spans), audio masking (none / remove), prosody-enhanced spans (on/off), and audio statistics pooling fusion (on/off), holding all other settings fixed. Table~\ref{ablation} presents the performance of the proposed model and its ablated configurations across four languages in the monolingual setting. The results reveal the relative contribution of each input modality, text, audio, prosody, and audio pooling statistics, to overall classification performance.

For Arabic, the full model (Text + Audio + Audio pooling stats + Prosody) achieves the highest performance, along with the configuration without prosody (Text + Audio + Audio pooling stats), as both reach the highest F1-score. This suggests that prosodic features contribute little when audio pooling statistics are present. While text-only performance was noticeably lower, models incorporating audio-based features showed general improvement, with pooling statistics yielding the most consistent gains. However, audio-only configurations show consistently poor results, indicating that acoustic input alone is insufficient for discourse relation classification. 

For English, the full model achieved the best performance, indicating that all feature types contribute to optimal results for this language. However, the text-only model performs nearly on par with the full system and achieves the highest precision, indicating strong reliance on textual information. Notably, removing either audio pooling statistics or prosody from the full configuration resulted in lower performance, highlighting the importance of both components. While prosodic features help maintain performance, adding audio alone leads to lower results, suggesting that raw acoustic input may introduce noise. This suggests that English relies more on prosody than on low-level audio features when textual signals are already strong. 

In contrast, French shows a different pattern, where the best performance is achieved without prosody. The configuration combining text, audio, and audio pooling statistics outperforms the full model that includes prosody. Additionally, the Text + Prosody model performs well, confirming that prosodic cues are informative in French. These results suggest that while prosody is useful, it may not combine effectively with other modalities in the full system.

In Spanish, the text-only model achieves the best overall performance, though text with prosody achieves the highest precision. Neither prosodic features nor audio pooling statistics provide measurable overall gains. These results suggest that acoustic input generally interferes with strong text-based predictions in Spanish, likely due to limited training data. Per-class F1 scores for language-specific settings are provided in Table~\ref{per-class3} (Appendix \ref{sec:appendix3}).

Overall, these results confirm that while text remains the most robust modality, multimodal integration can enhance performance when tailored to each language.

\section{Conclusion}

We introduce an adapted method for constructing multilingual and multimodal datasets for implicit discourse relations in distantly related languages. We apply this method to construct multimodal (text-audio) datasets for implicit discourse relations in English, French, and Spanish. We propose a multimodal modeling approach that leverages both textual and acoustic information for implicit discourse relation classification. Comprehensive experiments across language-specific and multilingual training settings show that while text-based models consistently outperform audio-based models, integrating both modalities can enhance performance. We also find that multilingual training can provide significant improvements for low-resource languages.
In future work, we aim to investigate whether the visual modality can also contribute, together with text and audio modalities, to improve the performance. Additionally, we plan to explore how to control multimodal fusion, investigating how much each modality should contribute to optimize performance across languages and relations.

\bibliography{tacl2021}
\bibliographystyle{acl_natbib}

\appendix

\section{Detailed Experimental Settings}
\label{sec:appendix0}

Table~\ref{hyperparams} presents the training hyperparameters used in our experiments.

\begin{table*}[t]
\resizebox{\textwidth}{!}{%
\centering
\small
\begin{tabular}{ll}
\toprule
\textbf{Hyperparameter} & \textbf{Value} \\
\midrule
\multicolumn{2}{l}{\textit{LoRA Configuration}} \\
Rank $r$ & 16 \\
Scaling $\alpha$ & 16 \\
Dropout & 0.05 \\
Target modules & \texttt{q\_proj, k\_proj, v\_proj, o\_proj, up\_proj, down\_proj, gate\_proj} \\
\midrule
\multicolumn{2}{l}{\textit{Training}} \\
Batch size & 1 \\
Gradient accumulation steps & 8 \\
Effective batch size & 8 \\
Epochs & 7 \\
Optimizer & AdamW \\
Weight decay & 0.05 \\
LR (backbone) & $5 \times 10^{-5}$ \\
LR (fusion/classifier head) & $5 \times 10^{-5}$ \\
LR (audio-stats head) & $5 \times 10^{-4}$ \\
LR scheduler & Cosine \\
Warmup ratio & 0.1 \\
Max gradient norm & 1.0 \\
\midrule
\multicolumn{2}{l}{\textit{Loss Function}} \\
Classification loss & Weighted cross-entropy \\
Label smoothing & 0.0 \\
$\lambda_{\mathrm{cls}}$ & 1.0 \\
$\lambda_{\mathrm{LM}}$ & 0.0 \\
$\lambda_{\mathrm{contr}}$ & 0.0 \\
\midrule
\multicolumn{2}{l}{\textit{Argument Pair Fusion}} \\
Architecture & Bidirectional cross-attention \\
Projection dimension & 512 \\
Attention heads & 4 \\
Temperature $\tau$ & 0.07 \\
\midrule
\multicolumn{2}{l}{\textit{Prosody Features}} \\
Feature dimension (per word) & 9 \\
Attention heads & 4 \\
Fusion scalar $\gamma$ & Learnable (init.\ 0.1) \\
Audio-stats residual scale & 0.1 \\
\bottomrule
\end{tabular}
}
\caption{Hyperparameters used in our proposed model for classification.}
\label{hyperparams}
\end{table*}

\section{Per-Class Results}
\label{sec:appendix1}

Table~\ref{per-class1} shows the baseline results for each implicit discourse relation (Cause-Effect, Contrast, Temporal, and Elaboration) across languages and modalities (text, audio, and text-audio).

\begin{table*}
\footnotesize
\centering
\begin{tabular}{c|c|c|c|c|c|c}

Language & Data & Models & Cause-Effect & Contrast & Temporal & Elaboration\\ \cline{7-7}
                   \hline
\multirow{6}{*}{Arabic} & \multirow{2}{*}{Audio} 
                   & Prosodic+LogReg & 0.23 & 0.37 & 0.19 & 0.32  \\ 
                   & & Whisper+NN & 0.37 & 0.34 & 0.19 & 0.34 \\ 
                   \cline{2-7}
                   & \multirow{2}{*}{Text} 
                   & TF-IDF+LogReg. & 0.37 & 0.53 & 0.33 & 0.38 \\ 
                   & & BERT & 0.41 & 0.48 & 0.35 & 0.42\\                  
                  \cline{2-7}   
                  & \multirow{2}{*}{Both} & 
                    Prosodic + TF-IDF+ LogReg.  & 0.39 & 0.49 & 0.34 & 0.45\\ 
                        & & BERT + Wav2vec2 & 0.45 & 0.47 & 0.29 & 0.44 \\ 
\hline
\multirow{6}{*}{English} & \multirow{2}{*}{Audio} 
                   & Prosodic+LogReg & 0.24 & 0.29 & 0.33 & 0.32 \\ 
                   & & Whisper+NN & 0.27 & 0.50 & 0.32 & 0.32 \\ 
                   \cline{2-7}
                   & \multirow{2}{*}{Text} 
                   & TF-IDF+LogReg. & 0.39 & 0.44 & 0.42 & 0.32\\ 
                   & & BERT & 0.48 & 0.68 & 0.51 & 0.45 \\                  
                  \cline{2-7}   
                  & \multirow{2}{*}{Both} & 
                    Prosodic + TF-IDF+ LogReg.  & 0.39 & 0.45 & 0.41 & 0.33 \\ 
                        & & BERT + Wav2vec2 & 0.49 & 0.63 & 0.48 & 0.45 \\ 
\hline
\multirow{6}{*}{French} & \multirow{2}{*}{Audio} 
                   & Prosodic+LogReg & 0.30 & 0.26 &0.20 & 0.14 \\ 
                   & & Whisper+NN & 0.23 & 0.43 & 0.07 &0.43  \\ 
                   \cline{2-7}
                   & \multirow{2}{*}{Text} 
                   & TF-IDF+LogReg. & 0.26 & 0.44 & 0.51 & 0.22 \\ 
                   & & BERT & 0.23 & 0.53 & 0.30 & 0.61  \\                  
                  \cline{2-7}   
                  & \multirow{2}{*}{Both} & 
                    Prosodic + TF-IDF+ LogReg.  & 0.28 & 0.43 & 0.26 & 0.51  \\ 
                        & & BERT + Wav2vec2 & 0.46 & 0.57 & 0.47  & 0.59 \\ 
\hline
\multirow{6}{*}{Spanish} & \multirow{2}{*}{Audio} 
                   & Prosodic+LogReg & 0.21 & 0.17 & 0.18 & 0.22  \\ 
                   & & Whisper+NN & 0.24 & 0.25 & 0.18 &0.31 \\ 
                   \cline{2-7}
                   & \multirow{2}{*}{Text} 
                   & TF-IDF+LogReg. & 0.29 & 0.38 & 0.22 & 0.42  \\ 
                   & & BETO & 0.38 & 0.14 & 0.00 & 0.42\\                  
                  \cline{2-7}   
                  & \multirow{2}{*}{Both} & Prosodic + TF-IDF+ LogReg. &
0.31& 0.31& 0.23& 0.36 \\ 
                        & & BETO + Wav2vec2 & 0.38 & 0.32 & 0.34  & 0.47  \\ 
\hline                   
\end{tabular}
%}
\caption{\label{per-class1} F1 scores (\%) for each discourse relation using baseline methods in IDR classification across different modalities (Text, Audio, and Text-Audio). Results are macro-averaged on the test set.}
\end{table*}

\label{sec:appendix2}

Table~\ref{per-class2} shows the results of our proposed model for each implicit discourse relation across languages and modalities (text, audio, and text-audio) in two settings:

\begin{table*}[h]
\small
\centering
\begin{tabular}{c|c|cccc|cccc}
{Language} & \multirow{3}{*}{Data} & \multicolumn{4}{c|}{Language-Specific Training} & \multicolumn{4}{c}{Multilingual Training} \\
\cline{1-1}\cline{3-10}
 Test& &  Cause-E. & Cont. & Temp. & Elab. & Cause-E. & Cont. & Temp. & Elab. \\
\hline

\multirow{3}{*}{Ar} & Audio 
    & 0.00 & 0.00 & 0.00 & 0.44  & 0.47 & 0.00 & 0.00 & 0.00\\ 
& Text 
    & 0.37 & 0.52 & 0.34 & 0.47 & 0.58 & 0.55 & 0.11 & 0.29\\ 
& Both 
    & 0.50 & 0.60 & 0.41 & 0.40 & 0.45 & 0.50 & 0.27 & 0.23\\ 
\hline
\multirow{3}{*}{En} & Audio 
    & 0.00 & 0.46 & 0.00 & 0.21 & 0.00 & 0.44 & 0.24 & 0.00\\ 
& Text 
    & 0.43 & 0.64 & 0.51 & 0.52 & 0.52 & 0.62 & 0.51 & 0.46\\ 
& Both 
    & 0.51 & 0.66 & 0.46 & 0.51  &0.50  &0.68 & 0.30 & 0.43 \\ 
\hline
\multirow{3}{*}{Fr} & Audio 
    & 0.00 & 0.37 & 0.00 & 0.07  & 0.00 & 0.00 & 0.00 & 0.00 \\ 
& Text 
    & 0.21 & 0.47 & 0.34 & 0.56  & 0.25 & 0.59 & 0.35 & 0.58\\ 
& Both 
    & 0.23 & 0.48 & 0.40 & 0.59  & 0.38 & 0.60 & 0.38 & 0.55 \\ 
\hline
\multirow{3}{*}{Es} & Audio 
    & 0.00 & 0.00 & 0.00 & 0.47  & 0.35 & 0.00 & 0.00 & 0.00\\ 
& Text 
    & 0.48 & 0.28 & 0.52 & 0.56  & 0.47 & 0.56 & 0.33 & 0.54\\ 
& Both 
    & 0.16 & 0.49 &  0.44 & 0.50  & 0.52 & 0.51 & 0.49 & 0.59 \\ 
\hline

\end{tabular}
%}
\caption{\label{per-class2} F1 scores (\%) for each discourse relation on the test set using our proposed IDR classification method across different modalities (Text, Audio, and Text-Audio). Performance shown for both monolingual training (models trained and tested on the same language) and multilingual training (models trained on all four languages and tested individually).}
\end{table*}

\label{sec:appendix3}

Table~\ref{per-class3} shows the results of the proposed model and its ablated configuration for each implicit discourse relation across four languages in the monolingual setting.

\begin{table*}[h]
\small
\footnotesize
\resizebox{\textwidth}{!}{%
\centering
\begin{tabular}{cccc|cccc|cccc|cccc|cccc}
\multicolumn{4}{c|}{Model Components} & \multicolumn{4}{c|}{Arabic} & \multicolumn{4}{c|}{English} & \multicolumn{4}{c|}{French} & \multicolumn{4}{c}{Spanish} \\
\hline
T & A & APS & Pr &Cause-E. & Cont. & Temp. & Elab. & Cause-E. & Cont. & Temp. & Elab. & Cause-E. & Cont. & Temp. & Elab. & Cause-E. & Cont. & Temp. & Elab. \\
\hline
\checkmark & \checkmark & \checkmark & \checkmark & 0.50 & 0.60 & 0.41 & 0.40 & 0.51 & 0.66 & 0.46 & 0.51 & 0.23 & 0.48 & 0.40 & 0.59 & 0.16 & 0.49 &  0.44 & 0.50 \\
\checkmark & \checkmark & \checkmark &   & 0.44 & 0.58 & 0.38 & 0.50 & 0.35 & 0.65 & 0.39 & 0.46 & 0.31 & 0.53 & 0.54 & 0.60 & 0.26 & 0.54 & 0.26 & 0.51 \\
\checkmark &   &   &   & 0.37 & 0.53 & 0.35 & 0.48 & 0.43 & 0.64 & 0.51 & 0.52 & 0.21 & 0.47 & 0.34 & 0.56 & 0.48 & 0.29 & 0.53 & 0.56 \\
\checkmark & \checkmark &   & \checkmark & 0.52 & 0.58 & 0.38 & 0.39 & 0.47 & 0.67 & 0.44 & 0.49 & 0.18 & 0.48 & 0.47 & 0.53 & 0.27 & 0.44 & 0.16 & 0.48 \\
\checkmark &   &   & \checkmark & 0.52 & 0.58 & 0.38 & 0.39 & 0.48 & 0.59 & 0.47 & 0.51 & 0.29 & 0.53 & 0.48 & 0.56 & 0.38 & 0.45 & 0.09 & 0.54 \\
\checkmark & \checkmark &   &   & 0.42 & 0.58 & 0.36 & 0.44 & 0.36 & 0.64 & 0.07 & 0.45 & 0.18 & 0.51 & 0.42 & 0.59 & 0.14 & 0.50 & 0.17 & 0.50 \\
  & \checkmark & \checkmark & \checkmark & 0.00 & 0.00 & 0.45 & 0.00 & 0.00 & 0.46 & 0.00 & 0.21 & 0.03 & 0.39 & 0.00 & 0.04 & 0.00 & 0.00 & 0.00 & 0.46 \\
  & \checkmark & \checkmark &   & 0.47 & 0.00 & 0.00 & 0.00 & 0.00 & 0.47 & 0.00 & 0.00 & 0.38 & 0.00 & 0.00 & 0.00 & 0.00 & 0.00 & 0.00 & 0.46 \\
\hline
\end{tabular}
}
\caption{\label{per-class3} F1 scores (\%) for each discourse relation in ablation study on the test set using our proposed IDR classification method across different modalities (Text, Audio, and Text-Audio). Performance shown for the monolingual setting (models trained and tested on the same language). T=Text, A=Audio, APS=Audio Pooling Stats, Pr=Prosody.}
\end{table*}

%\fi

\end{document}